\pdfoutput=1
\documentclass{article}

\usepackage{arxiv}

\usepackage[utf8]{inputenc} 

\usepackage{float}  

\usepackage{microtype}
\usepackage{graphicx}
\usepackage{subfigure}
\usepackage{booktabs} 
\usepackage{rotating} 
\usepackage{amsmath}
\usepackage{amssymb}
\usepackage[colorlinks=true, allcolors=blue]{hyperref}
\usepackage{multirow}
\usepackage{algorithm2e}
\RestyleAlgo{ruled}

\title{PSO-PINN: Physics-Informed Neural Networks Trained with Particle Swarm Optimization}

\author{
  Caio Davi \hspace{0.5cm} Ulisses Braga Neto\\
  Department of Electrical and Computer Engineering\\
  Texas A\&M University \\
  College Station, TX 77843 USA \\
  \texttt{\{caio.davi,ulisses\}@tamu.edu}
}

\begin{document}
\maketitle

\begin{abstract}

Physics-informed neural networks (PINN) have recently emerged as a promising application of deep learning in a wide range of engineering and scientific problems based on partial differential equation (PDE) models. However, evidence shows that PINN training by gradient descent displays pathologies that often prevent convergence when solving PDEs with irregular solutions. In this paper, we propose the use of a particle swarm optimization (PSO) approach to train PINNs. The resulting PSO-PINN algorithm not only mitigates the undesired behaviors of PINNs trained with standard gradient descent but also presents an ensemble approach to PINN that affords the possibility of robust predictions with quantified uncertainty. \textcolor{black}{We also propose PSO-BP-CD (PSO with Back-Propagation and Coefficient Decay), a hybrid PSO variant that combines swarm optimization with gradient descent, putting more weight on the latter as training progresses and the swarm zeros in on a good local optimum. Comprehensive experimental results show that PSO-PINN with the proposed PSO-BP-CD algorithm outperforms PINN ensembles trained with other PSO variants or with pure gradient descent.} 
\end{abstract}

\section{Introduction}
\label{intro}

Physics-informed machine learning is an emerging area that promises to have a  lasting impact in science and engineering. 
Physics-informed neural networks (PINNs)~\cite{raissi2019physics}, in particular, have shown remarkable success in a variety of problems modeled by partial differential equations~\cite{raissi2020hidden,karniadakis2021physics}. PINNs 
exploit the universal approximation capabilities of neural networks \cite{hornik1989multilayer}. Unlike traditional neural networks, PINNs employ a multi-part loss function containing data-fitting and PDE residual components. And differently from traditional numerical methods, PINNs are meshless, producing a solution over an entire, possibly irregular, domain, rather than on a pre-specified grid of points.

Standard gradient descent tools already in widespread use for training deep neural networks, such as stochastic gradient~\cite{amari1993backpropagation} and ADAM~\cite{kingma2014adam}, were promptly adopted as the methods of choice to train PINN. However, an accumulating body of evidence shows that gradient descent exhibits pathological behavior when training PINNs, especially when the differential equation solution contains irregular and fast varying features~\cite{wang2021understanding,wang2022and}. Numerous approaches have been proposed to mitigate this undesirable behavior, including weighting the loss function~\cite{wang2022and,mcclenny2020self, van2021optimally}, 
domain decomposition\cite{jagtap2020extended, shukla2021parallel}, and changes to the neural network architecture~\cite{wang2021understanding}.
Recent work attributed this behavior to  stiffness in the gradient flow dynamics of the PINNs loss functions, which leads to unstable convergence for gradient-based optimization algorithms~\cite{wang2021understanding, wang2022and}. 

In this work, we propose to address this issue by moving away from pure gradient descent by employing particle swarm optimization (PSO)~\cite{kennedy1995particle},  a metaheuristic algorithm used in numerous applications, including neural network training~\cite{chakraborty2017swarm, van2000cooperative, das2014artificial, mirjalili2015effective, mousavirad2020benchmark}. Indeed, the original paper on PSO~\cite{kennedy1995particle} was motivated in part by neural network training. The resulting PSO-PINN algorithm not only mitigates the pathologies associated with grandient-based optimization, but also produces a diverse ensemble of neural networks~\cite{hansen1990neural}, which enables the application of ensemble methods to PINNs, such as variance reduction\cite{krizhevsky2012imagenet} and uncertainty quantification\cite{lakshminarayanan2016simple}. Indeed, ensemble methods have become popular in deep learning as they combine the outputs of a diverse set of neural networks, thus reducing prediction variance and producing uncertainty estimates~\cite{deng2014ensemble,qiu2014ensemble,cao2020ensemble,ganaie2021ensemble}. To our knowledge, PSO-PINN is the first algorithm for training PINNs based on swarm optimization. However, PSO-PINN is not the only ensemble approach to PINNs; the ensemble approach was used to progressively train PINNs forward in time in a recent publication~\cite{haitsiukevich2022improved}.

In this paper, we consider and contrast three variants of the PSO algorithm to train PINNs: \textcolor{black}{the PSO method in its original form~\cite{kennedy1995particle}, a variant called PSO-BP (PSO with Back-Propagation), which adds a gradient descent component to the particle velocity vectors~\cite{yadav2019ga}, and PSO-BP-CD (PSO-BP with Coefficient Decay), a proposed modification of PSO-BP that includes a decreasing schedule for the behavioral coefficients. PSO-BP-CD puts more weight on the gradient descent component near the end of training, when the swarm should have already converged on a good local optimum.} Experimental results with several ODE and PDE benchmarks show that PSO-PINN, using the proposed PSO-BP-CD algorithm, consistently outperforms the other PSO variants for training PINNs, as well as PINN ensembles trained with standard ADAM~\cite{kingma2014adam}, in diverse scenarios under various parameter settings. These results demonstrate the potential of PSO-PINN in providing accurate prediction with quantified uncertainty.

This paper is organized as follows. Section \ref{sec:background} provides a brief overview of deep neural networks, physics-informed neural networks, and particle swarm optimization. Section \ref{sec:method} introduces the PSO-PINN algorithm based on three variants of PSO, plain, PSO-BP, and PSO-BP-CD, whereas Section \ref{sec:results} studies PSO-PINN through experimental results using various ODE and PDE benchmarks. Section \ref{sec:conclusion} provides concluding remarks and future directions.

\def\bx{\mathbf{x}}
\def\by{\mathbf{y}}

\section{Background}
\label{sec:background}

In this section, we define PINNs and describe the PSO-BP optimization algorithm. 

\subsection{Deep Neural Networks}

The feed-forward fully-connected neural network is the basic architecture used in deep learning algorithms\cite{lecun2015deep}. A fully-connected neural network with $L$ layers is a function $f_\theta:\mathbb{R}^d \rightarrow \mathbb{R}^k$ described by
\begin{equation}
    \label{eq:fully-connectedNN}
    f_\theta(\bx) \,=\, W^{[L-1]}\sigma \circ 
                ( \cdots
                \sigma \circ (W^{[0]}\bx + b^{[0]}) +                 \cdots) 
                + b^{[L-1]}
\end{equation}
where $\sigma$ is an entry-wise activation function, $W^{[l]}$ and $b^{[l]}$ are respectively the weight matrices and the bias corresponding to each layer $l$, and $\theta$ is the set of weights and biases:
\begin{equation}
\label{eq:theta}
    \theta = (W^{[0]}, \cdots ,
    W^{[L-1]}, b^{[0]}, \cdots, b^{[L-1]})\,.
\end{equation}
Among popular choices for the activation function are the sigmoid function, the hyperbolic tangent function (tanh), and the rectified linear unit (ReLU)~\cite{glorot2011deep}. To fit the neural network $f_\theta(\bx)$ to  data $\{(\bx_i,\by_i)\}_{i=1}^n$, one minimizes a suitable loss function. A popular choice in traditional deep learning is the mean square error (MSE): 
\begin{equation}
    \label{eq:loss}
    \mathcal{L} \,=\, \frac{1}{n} \sum_{i=1}^n ||f_\theta (\bx_i) - \by_i||^2.
\end{equation}

\subsection{Physics-Informed Neural Networks}
\label{sec:PINN}

Consider a non-linear differential equation of the general form:
\begin{equation}
\begin{split}
    \mathcal{N}[u(\mathbf{x})] & \,=\, g(\mathbf{x})\,, \quad \mathbf{x} \in \Omega\,, \\
    u(\mathbf{x}) &\,=\, h(\mathbf{x})\,, \quad \mathbf{x} \in \partial\Omega\,,
\end{split}    
\end{equation}
where $\Omega \subset \mathbb{R}^{d}$, $u: \Omega \rightarrow \mathbb{R}^k$, $\mathcal{N}[\cdot]$ is a differential operator, and $g,h: \Omega \rightarrow \mathbb{R}^k$ specify forcing and boundary conditions, respectively. We employ a neural network $f_\theta(\bx)$ to approximate the unknown function $u(\bx)$. The PDE is enforced through the loss function
\begin{equation}
\label{eq:residual_loss}
        \mathcal{L}_r \,=\,
        \frac{1}{n_r}\sum_{i=1}^{n_r} ||\mathcal{N}[f_\theta(\mathbf{x}^r_i)]-g(\mathbf{x}^r_i)||^2,
\end{equation}
where $\{\mathbf{x}^r_i\}_{i=1}^{n_r} \subset \Omega$ are collocation points, and $\mathcal{N}[f_\theta(\bx)]$ is computed accurately using 
automatic differentiation methods~\cite{baydin2018automatic}.
The neural network is fitted to the initial and boundary condition $h(\bx)$ using a traditional data-fitting loss function:
\begin{equation}
\label{eq:boundary_loss}
    \mathcal{L}_b = 
    \frac{1}{n_b}\sum_{i=1}^{n_b} ||f_\theta(\mathbf{x}^b_i)-h(\mathbf{x}^b_i)||^2
\end{equation}
where $\{\mathbf{x}^b_i\}^{n_b}_{i=1} \subset \partial \Omega$ are initial and boundary points. If there are experimental or simulation data $\{(\mathbf{x}^d_i, \mathbf{y}^d_i)\}^{n_d}_{i=1}$ available, they may be included in the usual way through the loss:
\begin{equation}
\label{eq:data_loss}
    \mathcal{L}_d =
    \frac{1}{n_d}\sum_{i=1}^{n_d}||f_\theta(\mathbf{x}^d_i)- \mathbf{y}^d_i||^2.
\end{equation}

The previous multi-objective optimization problem is typically solved by simple linear scalarization, in which case the PINN is trained by minimizing the total loss function $\mathcal{L}$:
\begin{equation}
    \label{eq:pinn_loss}
    \mathcal{L} \,=\, \mathcal{L}_r + \mathcal{L}_b + \mathcal{L}_d\,.
\end{equation}

This framework is easily extended to more complex initial and boundary conditions involving derivatives of $u(\bx)$. 

\subsection{Particle Swarm Optimization}
\label{sec:pso}

\begin{figure}[t]
    \begin{center}
    \includegraphics[width=0.8\columnwidth]{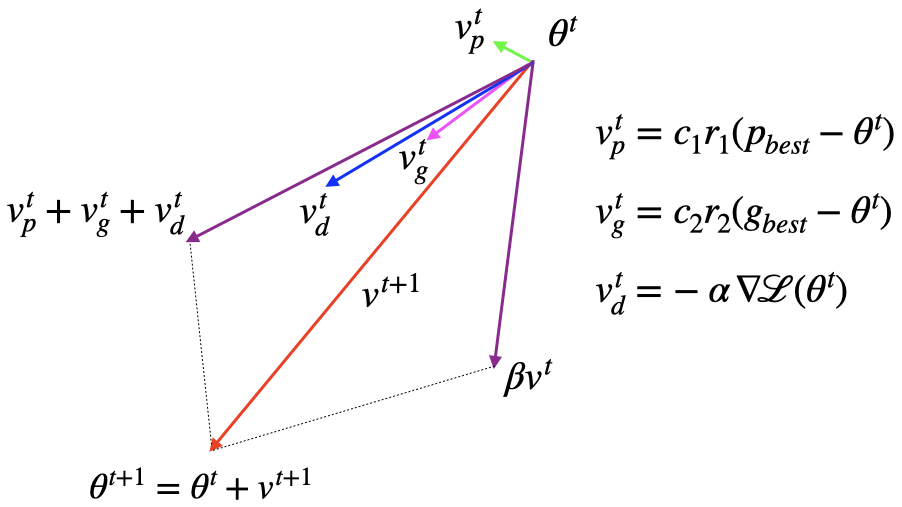}    
    \hfill
    \end{center}
    \caption{
    %
    The PSO-BP update.
    }
    \label{fig:pso-pinn}
  \end{figure}

The Particle Swarm Optimization (PSO) algorithm~\cite{eberhart1995particle, shi1999empirical} is a population-based stochastic optimization algorithm that emulates the swarm behavior of particles distributed in a $n$-dimensional search space~\cite{wang2018particle}. Each individual in this swarm represents a candidate solution. At each iteration, the particles in the swarm exchange information and use it to update their positions. Particle $\theta^t$ at iteration $t$ is guided by a velocity determined by three factors: its own velocity inertia $\beta v^t$, its best-known position $p_{best}$ in the search-space, as well as the entire swarm's best-known position $g_{best}$:
\begin{equation}
    \label{eq:pso-velocity}
    v^{t+1} \,=\, \beta v^t + c_1r_1(p_{best}-\theta^t) + c_2r_2(g_{best}-\theta^t)\,,
\end{equation}
where $c_1$ and $c_2$ are the cognitive and social coefficients, respectively, referred to jointly as the behavioral coefficients, and $r_1$ and $r_2$ are uniformly distributed random numbers in range $[0,1)$.
Then the particle position is updated as
\begin{equation}
    \label{eq:pso-location}
    \theta^{t+1} \,=\, \theta^t + v^{t+1}\,.
\end{equation}
Many variations of the PSO algorithm were proposed over time. PSO can be combined with gradient descent to train neural networks, with the gradients computed efficiently by automatic differentiation (backpropagation), as was done in \cite{yadav2019ga}. In this  algorithm, which was called PSO-BP by the authors, the  particle velocity has an additional component, namely, the gradient vector of the loss function$\nabla \mathcal{L}(\theta)$:
\begin{equation}
    \label{eq:pso-bp-velocity}
    v^{t+1} \,=\, \beta v^t + c_1r_1(p_{best}-\theta^t) + c_2r_2(g_{best}-\theta^t) - \alpha \nabla \mathcal{L}(\theta^t)\,,
\end{equation}
where $\alpha$ is a learning rate. Therefore, the gradient participates in the velocity magnitude and direction, by an amount that is specified by the learning rate, \textcolor{black}{which helps guide the swarm to a good local optimum}. See Figure~\ref{fig:pso-pinn} for an illustration of the PSO-BP update.

\section{PSO-PINN Algorithm}
\label{sec:method}

In this section, we describe three variants of the PSO algorithm that we use to train PINNs, how to initialize them, and a few possible ways to summarize the ensemble results.

\subsection{The PSO variants}
\label{sec:models}

We consider the following three variants of the PSO algorithm:

\begin{enumerate}
    \item{
        \textbf{PSO} - The algorithm in its original form, as described in \eqref{eq:pso-velocity} and \eqref{eq:pso-location}.

    }
    \item{
        \textbf{PSO-BP} - This version adds a component based on the gradient of the loss function, as described in \eqref{eq:pso-bp-velocity}.
    }
    \item{
        \textbf{PSO-BP-CD}
        In this variant, we introduce a behavioral coefficient schedule. The behavioral components $c_1$ and $c_2$ in PSO-BP-CD decay linearly according to the iteration number as follows:
        \begin{equation}
            c^{t+1}_k = c^t_k - \frac{2c^t_k}{n}
        \end{equation}
        for $k={1,2}$ and $t={1,2,\ldots,n}$, where $n$ is the total number of iterations.
    }

\end{enumerate}

\textcolor{black}{We propose PSO-BP-CD as a variant of PSO-BP, which puts more weight on the gradient vector near the end of training, when the swarm should already have zeroed in on a good local optimum, and less communication between the particles is needed. This allows PSO-BP-CD to be more effective than the PSO-BP and pure PSO methods in training PINN ensembles, as will be seen in the experimental results section.}

\subsection{The Algorithm}

The PSO-PINN swarm is an ensemble of PINN candidates. Each particle in the swarm is a vector $\theta$ containing the weights and bias of the corresponding PINN.
The population $\Theta$ of all particles is randomly initialized, as will be detailed later in Section~\ref{sec:model_selection}. The objective function $f_\theta(\cdot)$ is the total loss function in~\eqref{eq:pinn_loss}. The $p_{best}$ vector should be initialized to the values in the original population $\Theta$, as the particles know nothing but their initial point at this time. The $g_{best}$ parameter is equal to the value of $\theta \in \Theta$ that achieves the minimum value of $f(\theta)$. After the initialization step, the updates to the swarm attempt to minimize the expected value of the objective function. Each particle velocity is updated following~\eqref{eq:pso-bp-velocity}. Despite the use of the gradient update $\alpha \nabla \mathcal{L}(\theta)$ in the previous equation, in practice, the update is made using ADAM optimization~\cite{kingma2014adam}. After the velocity update, the particle locations (i.e., the network weights) are updated according to \eqref{eq:pso-location}. The values of $p_{best}$ and $g_{best}$ are updated accordingly. The training loop runs until a maximum number of iterations is reached. The procedure is summarized in Algorithm~\ref{alg:PSO-GD}.

\begin{algorithm}[t!]
\caption{PSO-PINN}\label{alg:PSO-GD}
\textbf{Require: } $\alpha$: step size;  \\
\textbf{Require: } $\beta$: inertia; \\
\textbf{Require: } $c_1$, $c_2$: behavioral coefficients;\\
\textbf{Require: } $\mathcal{L}$: total loss function for the PINN; \\
Initialize population $\Theta$; \\
$p_{best}(i) \gets \theta_i$, for $i=1,\ldots,|\Theta|$; \\ 
$g_{best} \gets \arg\min_{\,\theta \in p_{best}} \mathcal{L}(\theta)$; \\ 
\textbf{for} $t = 1,2,\ldots,$ MAX \textbf{do}:\\
\text{\quad}\textbf{for}  $i=1,\ldots,|\Theta|$ \textbf{do}: \\ 
  \text{\quad}\text{\quad}$r_1, r_2 \gets U(0,1]$; \\
  \text{\quad}\text{\quad}$V \,=\, \beta V \,+ \, c_1r_1(p_{best}(i)-\theta_i) \,+\, c_2r_2(g_{best}-\theta_i)$ \\
  \text{\quad}\text{\quad}\textbf{if} $BP$:\\
    \text{\quad}\text{\quad}\text{\quad}$V \,=\, V \,+ \, \alpha \nabla \mathcal{L}(\theta_i)$\\ 
  \text{\quad}\text{\quad}$\theta_i \,=\, \theta_i \,+\, V$; \\
  \text{\quad\quad}\textbf{if} $\mathcal{L}(\theta_i) < \mathcal{L}(p_{best}(i))$:\\
  \text{\quad\quad\quad}$p_{best}(i) \gets \theta_i$\\
\text{\quad}\textbf{end}\\
  \text{\quad}$g_{best} \gets \arg\min_{\,\theta \in p_{best}}\mathcal{L}(\theta)$; \\
  \text{\quad}\textbf{if} {\em coefficient\_decay}:\\
  \text{\quad}\text{\quad}$c_1 \,=\, c_1 \,-\,\frac{2c_1}{t}$;\\
  \text{\quad}\text{\quad}$c_2 \,=\, c_2 \,-\,  \frac{c_2}{t}$;\\
\textbf{end}
\end{algorithm}

\subsection{The Ensemble Solution}

Besides the fact that swarms are a well-established optimization method, they also provide an ensemble of solutions. Thus, we can take advantage of the various properties provided by ensembles. This category of learning methods combines several individual models to create a ``collective'' solution, which is expected to display improved performance and stability with respect to any of the individual models.  
There are different methods to seek the consensus decision in an ensemble~\cite{ganaie2021ensemble}. The most straightforward one perhaps would be simply choosing the best-fitted model according to the loss function. Although this may yield a good model, it could fail due to overfitting. To avoid this, a better approach exploits the ensemble diversity through model averaging: 
\begin{equation}
    \hat{f}(\mathbf{x}) = \frac{1}{|\Theta|} \sum_{\theta_i \in \Theta} f_{\theta_i}(\mathbf{x})\,.
\end{equation}
Uncertainty of the prediction at each point $\mathbf{x}$ of the domain can be quantified naturally through the sample variance:
\begin{equation}
    \hat{\sigma}^2(\mathbf{x}) = \frac{1}{|\Theta| - 1}\,
    \sum_{\theta_i \in \Theta} (f_{\theta_i}(\mathbf{x}) - \hat{f}(\mathbf{x}))^2.
\end{equation}
\section{Experimental Results}
\label{sec:results}

In this section, we study the performance of PSO-PINN empirically, using the three PSO variants discussed earlier, by means of several classical ODE and PDE benchmarks. In addition,  PSO-PINN ensembles are compared to traditional ensembles of PINNs trained with ADAM gradient descent. All experiments employ fully-connected architectures, with the hyperbolic tangent activation function and Glorot initialization of the weights~\cite{glorot2010understanding}. 
The main figure of merit used is the $L_2$ error:
\begin{equation}
    L_2 \ \text{error} \,=\, \frac{\sqrt{\sum_{i=1}^{N_U} |\hat{f}(\mathbf{x}_i) - U(\mathbf{x}_i)|^2}}{\sqrt{\sum_{i=1}^{N_U} |U(\mathbf{x}_i) |^2}}\,.
\end{equation}
where $\hat{f}(\mathbf{x})$ is the prediction and $U(\mathbf{x})$ is the analytical solution or a high-fidelity approximation over a test mesh $\{\mathbf{x}_i\}_{i=1}^{N_U}$. All experiments in this section were performed using Tensorflow 2~\cite{tensorflow2015-whitepaper}.


\subsection{A Simple PSO-PINN Example}

Fist, we use the 1D Poisson equation to illustrate the PSO-PINN algorithm. Due to its simplicity, this example allows us to visualize the evolution of training and the results. The 1D Poisson equation considered here is:
\begin{equation}
    \begin{split}
        &u_{xx} = g(x)\,, \quad x \in [0,1]\,,  \\ 
        &u(0) = u(1) = 0\,.
    \end{split}
\end{equation}
where $g(x) = -(2\pi)^{2}\sin(2\pi x)$ is manufactured so that the solution is $u(x) = \sin(2\pi x)$. For this simple problem, we use a relatively shallow architecture consisting of 3 layers of 10 neurons. The PSO-PINN ensemble was composed by 50 particles and used the proposed PSO-BP-CD optimization method,
with hyperparameter values $\alpha=0.005$, $\beta=0.9$, $c_1=0.08$, and $c_2=0.5$. 

\begin{figure*}[t]
    \begin{center}
    \includegraphics[width=1\columnwidth]{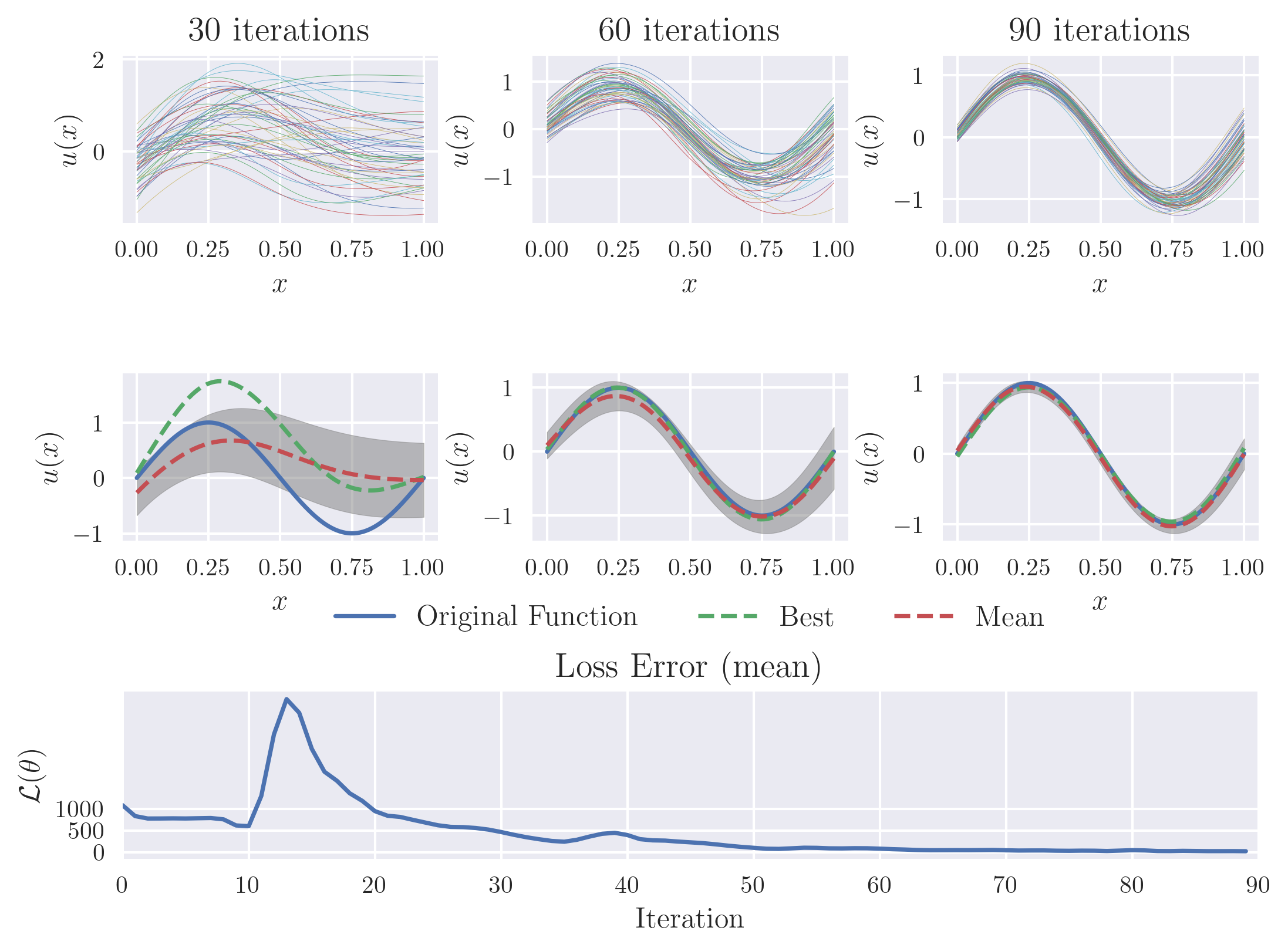}    
    \hfill
    \end{center}
    \caption{
    %
    \textbf{Poisson Equation} - First row: solutions in the PSO-PINN ensemble as training progresses, using the PSO-BP-CD method. Second row: Best individual, mean and variance of PSO-PINN ensemble. Third row: PSO-PINN mean loss error. The grey band is the two-sided 1-standard deviation region.}
    \label{fig:poisson_train}
\end{figure*}

We can see in Figure \ref{fig:poisson_train} the progression of training, and how the PSO-PINN swarm zeros in on the solution, rather quickly, after only 90 iterations. We can observe that, initially, the mean is off and the variance is large, which is the desired behavior, as the variance is supposed to quantify the uncertainty associated with the approximation. As training progresses, the swarm approaches a consensus, the mean converges to the solution, and the variance simultaneously shrinks to zero, indicating more confidence in the final result. The PSO-PINN swarm is thus accurate, with well-calibrated uncertainty quantification.


\subsection{PSO-PINN Performance using Different PSO Variants}
\label{sec:model_selection}

In this section, we use classical PDE benchmarks to study the performance of PSO-PINN using the three PSO variants described in \ref{sec:models}.  The number of training iterations was fixed at 2,000, while the architecture of the neural networks was fixed at 5 layers and 8 neurons per layer. The ensemble size was kept fixed at 100 neural networks. We kept the same number of initial condition points, boundary condition points and residual points for all experiments in this section. In all experiments, we used 1024 evenly spaced points for the initial condition, 512 evenly spaced points for the boundary conditions, 
and 1000 residual points randomly distributed over the solution domain using latin hypercube sampling. For the basic PSO algorithm, the hyperparameters were set to $\beta=0.9$, $c_1=0.8$ and $c_2=0.5$ in all experiments.



\subsubsection{Advection Equation}

The advection equation models the transport of a substance by bulk motion of a fluid. In this example, we are assuming a linear univariate advection equation~\cite{leveque2002finite}:
    
    \begin{equation}
      q_t + uq_x = 0
      \end{equation}
    where $u$ is the constant velocity. The Riemman initial condition is  
    \begin{equation}
         q(x,0) = 
            \begin{cases} 
                q_l, & 0 \leq x < x_0 \,, \\
                q_r , & x_0 < x \leq L\,. 
            \end{cases}
      \end{equation}
    This simple problem has as solution:
    \begin{equation}
         q(x,t) = 
            \begin{cases} 
                q_l, & 0 \leq x < x_0+ut \,, \\
                q_r, & x_0+ut < x \leq L\,,     
            \end{cases}
    \end{equation} 
    for $0 \leq t < (L-x_0)/u$. In other words, the initial discontinuity in concentration is simply advected to the left with constant speed $u$.

    \begin{figure}
    \begin{center}
    \includegraphics[width=1\columnwidth]{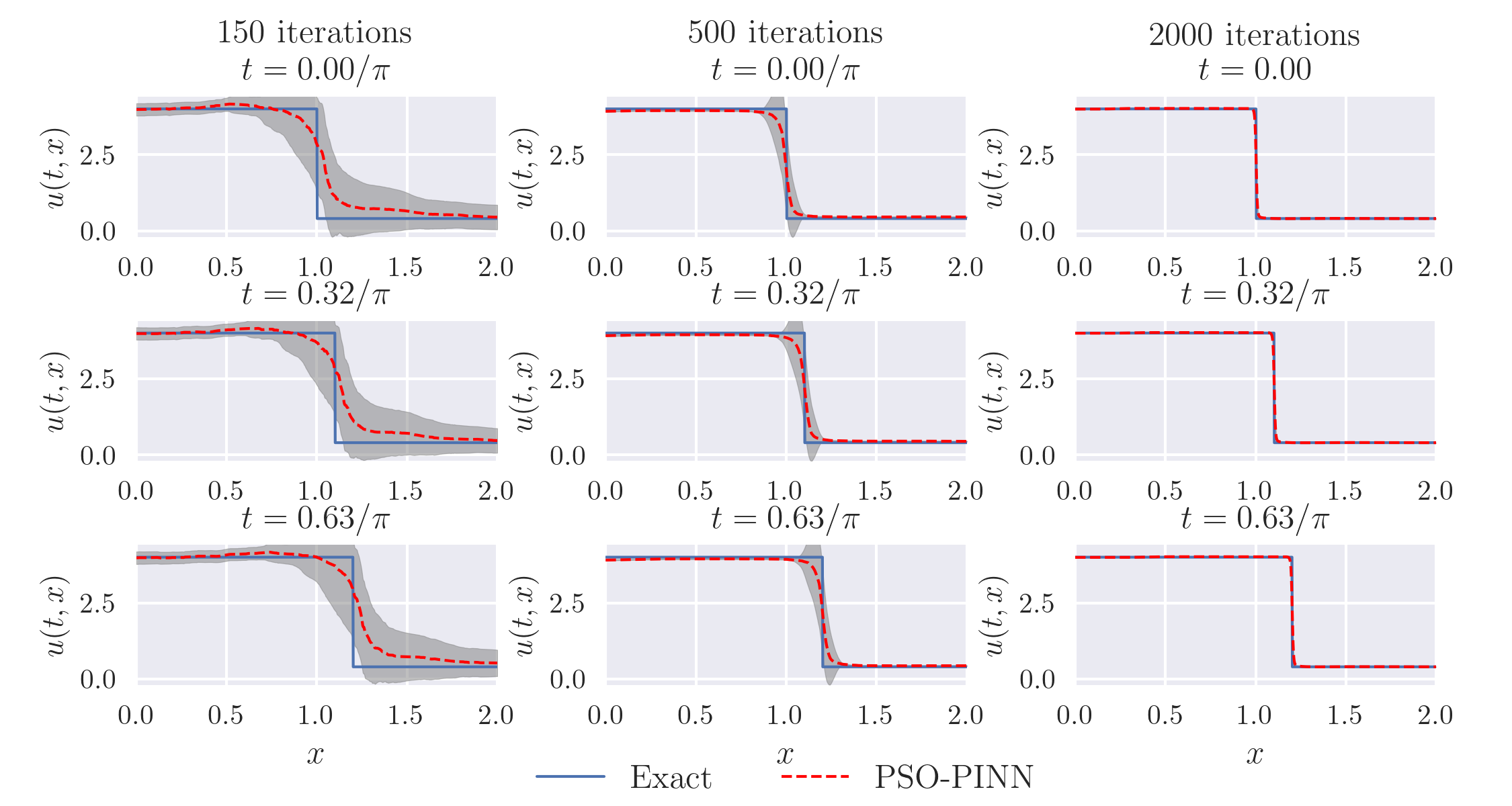}  
    \hfill
    \end{center}
    \caption{
    %
    \textbf{\bf Advection Equation} - Evolution of the PSO-PINN ensemble using the PSO-BP-CD method, showing the mean and two-sided 1-standard deviation of the ensemble.}
    \label{fig:advection}
    \end{figure}
    
    For both the PSO-BP and the PSO-BP-CD methods, the hyperparameters were set to $\alpha = 0.005$, $\beta=0.99$, $c_1=0.08$, and $c_2=0.5$ were used.
        Table~\ref{tab:PSO_variants} displays the performance of PSO-PINN using the three PSO variants, which shows that the PSO-BP-CD method produced the best average $L_2$ error over 10 independent repetitions of the experiment. 
        Figure~\ref{fig:advection} displays the evolution of training of the PSO-PINN ensemble using the PSO-BP-CD method, in one of the 10 tests. We can observe that the PSO-PINN ensemble shows a reasonable solution after 500 iterations, and has converged at 2000 iterations. As expected, the variance of the ensemble is largest when the approximation is off and it shrinks to nearly zero upon convergence to the analytical solution. 
    
      \subsubsection{Heat Equation} 
        
        Next, we consider the classical 1D Heat equation. This PDE models temperature dissipation in a heat-conducting  bar:
            \begin{equation}
            \label{eq:heat}
                \begin{split}
                    &\frac{\partial u}{\partial t}=\alpha \frac{\partial^2u}{\partial x^2}, \qquad
                    x \in [0, L], \quad t \in [0, 1] \\
                    &u(0,t) = u(L,t)=0,\\
                    &u(x,0) = \sin \Big(\frac{\pi x}{L}\Big),\qquad 0<x<L\,, \\
                \end{split}
            \end{equation}
        where $L=1$ is the length of the bar. The reference solution is $u(x,t) = e^{\frac{\pi ^2 \alpha t}{L^2}}\sin (\frac{\pi x}{L})$.
        
\begin{figure*}[t!]
            \begin{center}
            \includegraphics[width=1\columnwidth]{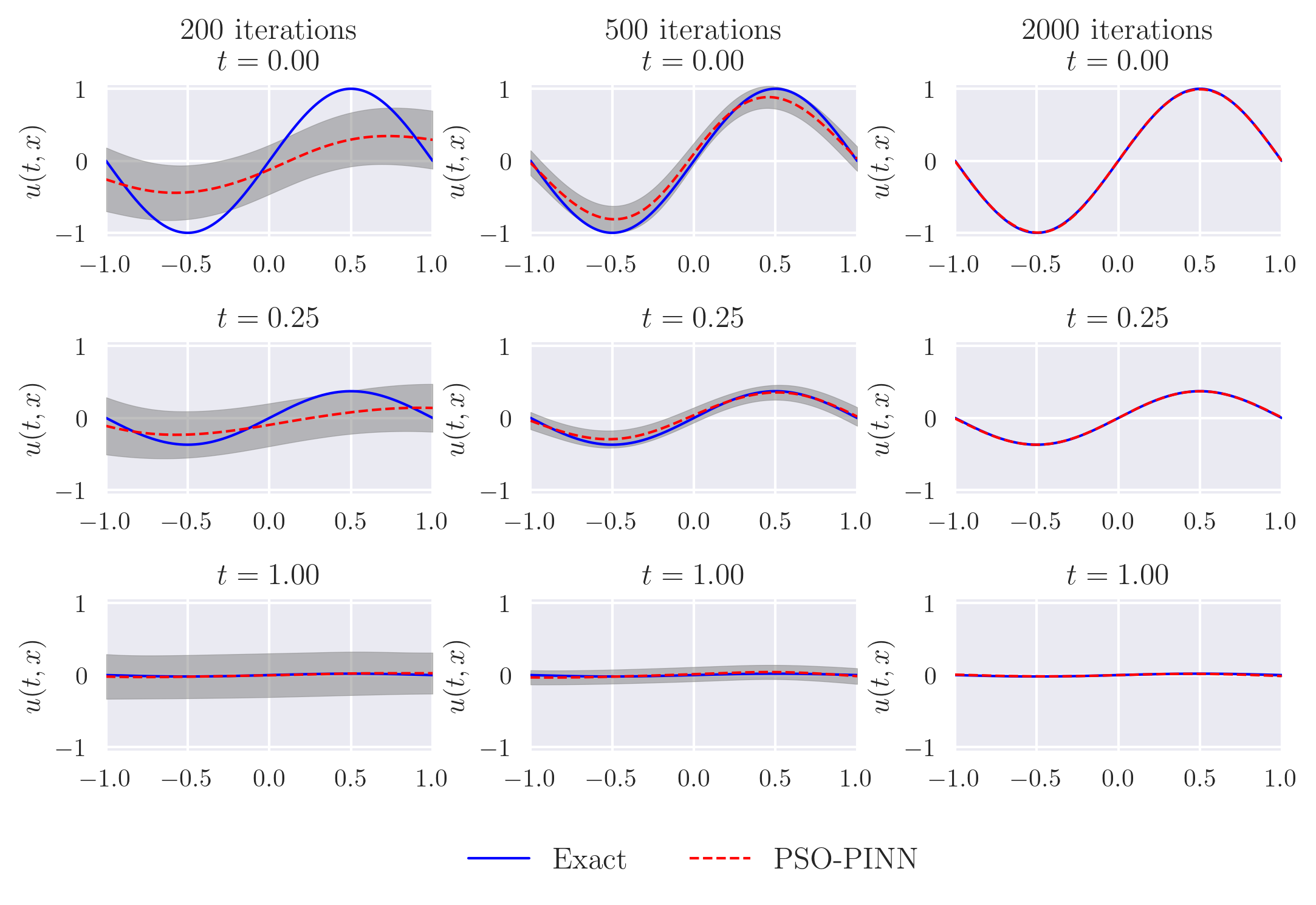}    
            \hfill
            \end{center}
            \caption{{\bf Heat Equation} - Evolution of the PSO-PINN ensemble using the PSO-BP-CD method, showing the mean and two-sided 1-standard deviation of the ensemble.}
            \label{fig:heat}
        \end{figure*}

      The hyperparameters used for PSO-BP and the PSO-BP-CD methods are the same as in the previous benchmark. The results are displayed in Table~\ref{tab:PSO_variants}, where again we can see that the PSO-BP-CD variant performed the best. Figure \ref{fig:heat} represents the training of the PSO-PINN for the heat equation. Figure~\ref{fig:burgers} illustrates the evolution of training for the PSO-BP-CD variant in one of the experiments. Once again, the PSO-PINN ensemble produces a reasonable solution after 500 iterations, has converged at 2,000 iterations, and displays larger variance when the approximation is farther from the reference solution.
        
     \subsubsection{Burgers Equation}
     
     Finally, we consider a nonlinear benchmark, namely, the well-known viscous Burgers equation with sinusoidal initial condition:
        \begin{equation}
        \label{eq:burgers}
            \begin{split}
                &\frac{\partial u}{\partial t} + u\frac{\partial u}{\partial x} = \nu\frac{\partial^2u}{\partial x^2}, \qquad x \in [-1, 1], \quad t \in [0, 1] \\
                &u(0,x) = -\sin(\pi x)\,, \\
                &u(t, -1) = u(t,1) = 0\,,
            \end{split}
        \end{equation}
        with kinematic viscosity $\nu = 0.02/\pi$. 
        
\begin{figure}
        \begin{center}
        \includegraphics[width=0.98\columnwidth]{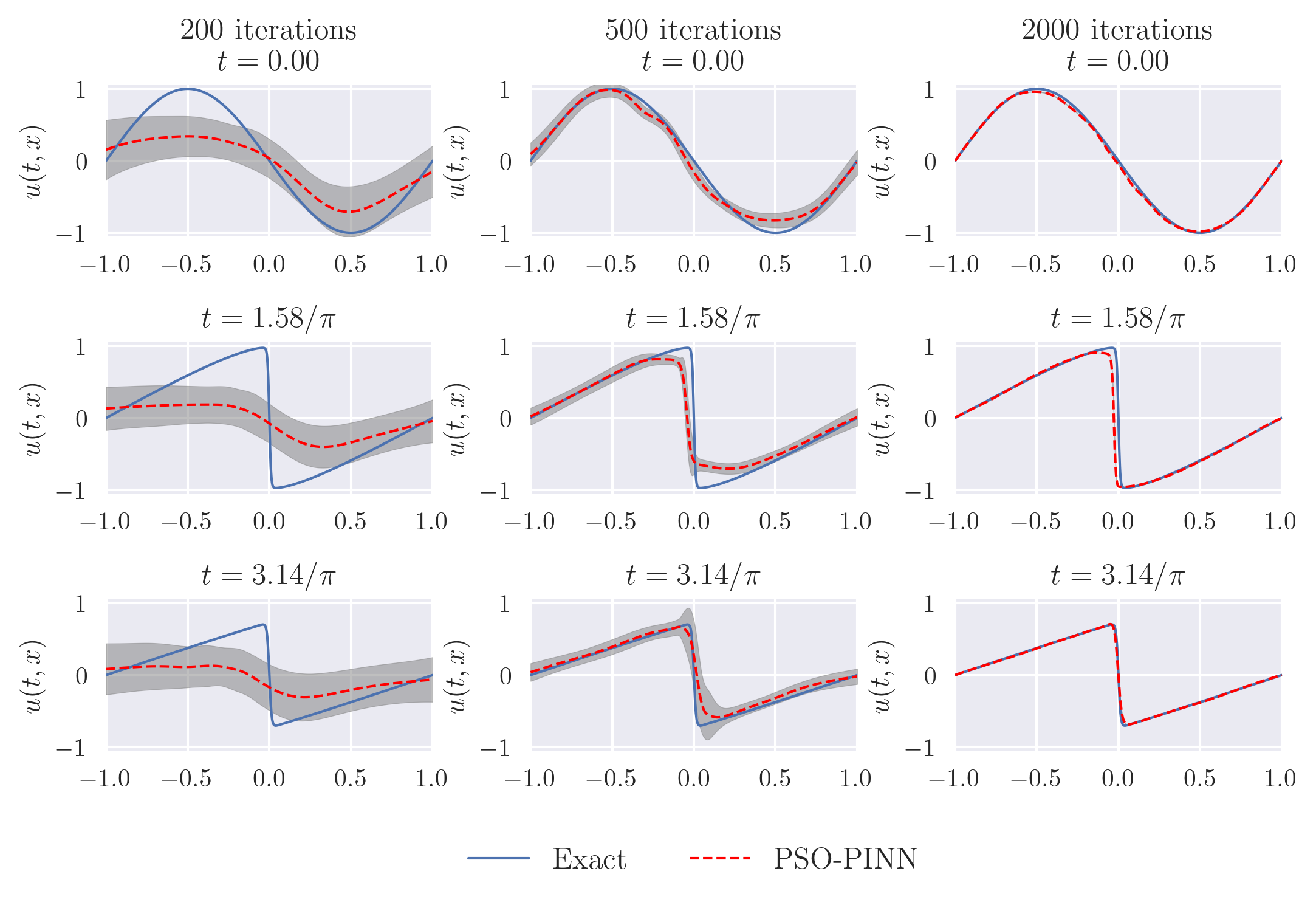}  
        \hfill
        \end{center}
        \caption{{\bf Burgers Equation} - Evolution of the PSO-PINN ensemble using the PSO-BP-CD method, showing the mean and two-sided 1-standard deviation of the ensemble.}
        %
        \label{fig:burgers}
        \end{figure}

         The nonlinearity of the PDE necessitated adjusting slightly the  hyperparameter values used in the PSO-BP and the PSO-BP-CD methods to $\alpha = 0.001$, $\beta=0.9$, $c_1=0.08$ and $c_2=0.05$.
        The results can be seen in Table~\ref{tab:PSO_variants}. It can be seen that the proposed PSO-BP-CD variant is better, by more than an order of magnitude, than the other variants. Figure~\ref{fig:burgers} illustrates the evolution of training for the PSO-BP-CD variant in one of the experiments. As in the case of the previous linear benchmarks, the PSO-PINN ensemble produces, for this much more complex PDE, a reasonable solution after 500 iterations, and has converged at 2,000 iterations. Note how the variance is larger when the approximation is farther from the reference solution.
        
\begin{table}[]
\centering
\caption{PSO-PINN results comparing the PSO variants, displaying the mean, standard deviation, and minimum (in parenthesis) of the L2 error of the PSO-PINN combined prediction across 10 independent repetitions of the experiment.}
\vspace{2ex}
\label{tab:PSO_variants}
\begin{tabular}{lccc}
\hline
Problem & PSO variant & $L_2$ error \\ \hline
\multirow{3}{*}{Advection} & PSO       & 0.1177$\,\pm\,$0.0714 \: (0.0391) \\
                           & PSO-BP    & 0.0234$\,\pm\,$0.0017 \: (\textbf{0.0212}) \\
                           & PSO-BP-CD & \textbf{0.0232}$\,\pm\,$0.0016 \: (0.0213) \\ \hline
\multirow{3}{*}{Heat}      & PSO       & 0.3974$\,\pm\,$0.0834 \: (0.2617) \\
                           & PSO-BP    & 0.0068$\,\pm\,$0.0020 \: (0.0027) \\
                           & PSO-BP-CD & \textbf{0.0051}$\,\pm\,$0.0026 \: (\textbf{0.0023}) \\ \hline
\multirow{3}{*}{Burgers}   & PSO       & 0.4197$\,\pm\,$0.0711 \: (0.2819) \\
                           & PSO-BP    & 0.1470$\,\pm\,$0.0826 \; (0.0579) \\
                           & PSO-BP-CD & \textbf{0.0125}$\,\pm\,$0.0502 \: (\textbf{0.0057})                           
\end{tabular}
\end{table}

\subsection{Comparison with Traditional Ensembles of PINNs}

In this section we show that PSO-PINN ensembles produce better results than  traditional ensembles of PINNs trained individually with ADAM gradient descent. 
For this, we selected two particularly hard examples, in order to appreciate the improvement over traditional ensembles afforded by PSO-PINN over traditional ensembles. In both benchmarks, the PSO-BP-CD variant is used with the PSO-PINN, with $\alpha = 0.005$, $\beta = 0.99$, $c_1=0.08$ and $c_2=0.5$, while the ADAM parameters were set to $\alpha=0.001$, $\beta_1=0.99$, and $\beta_2=0.999$.

\subsubsection{Forced Heat Equation} 
\label{sec:heat_}
            
This benchmark features a 1D forced heat equation:
    \begin{equation}
    \label{eq:diffusion}
        \begin{split}
            &\frac{\partial u}{\partial t}- \frac{\partial^2u}{\partial x^2} \,=\, 1 + x \cos(t) \quad
            x \in [-1, 1], \quad t \in [0, 1] \\
            &\frac{\partial u}{\partial x}\bigg\rvert_{x=0} = \frac{\partial u}{\partial x}\bigg\rvert_{x=1}=\, \sin(t),\\
            &u(x,0) = 1 + \cos(2 \pi x)\,. \\
        \end{split}
    \end{equation}
This problem has an exact solution, given by
\begin{equation}
 u(x,t) \,=\, 1 + t + e^{-4 \pi^2 t} \cos(2\pi x) + x \sin(t)\,.
\end{equation}
     
\begin{figure*}[t!]
    \begin{center}
        \includegraphics[width=0.9\columnwidth]{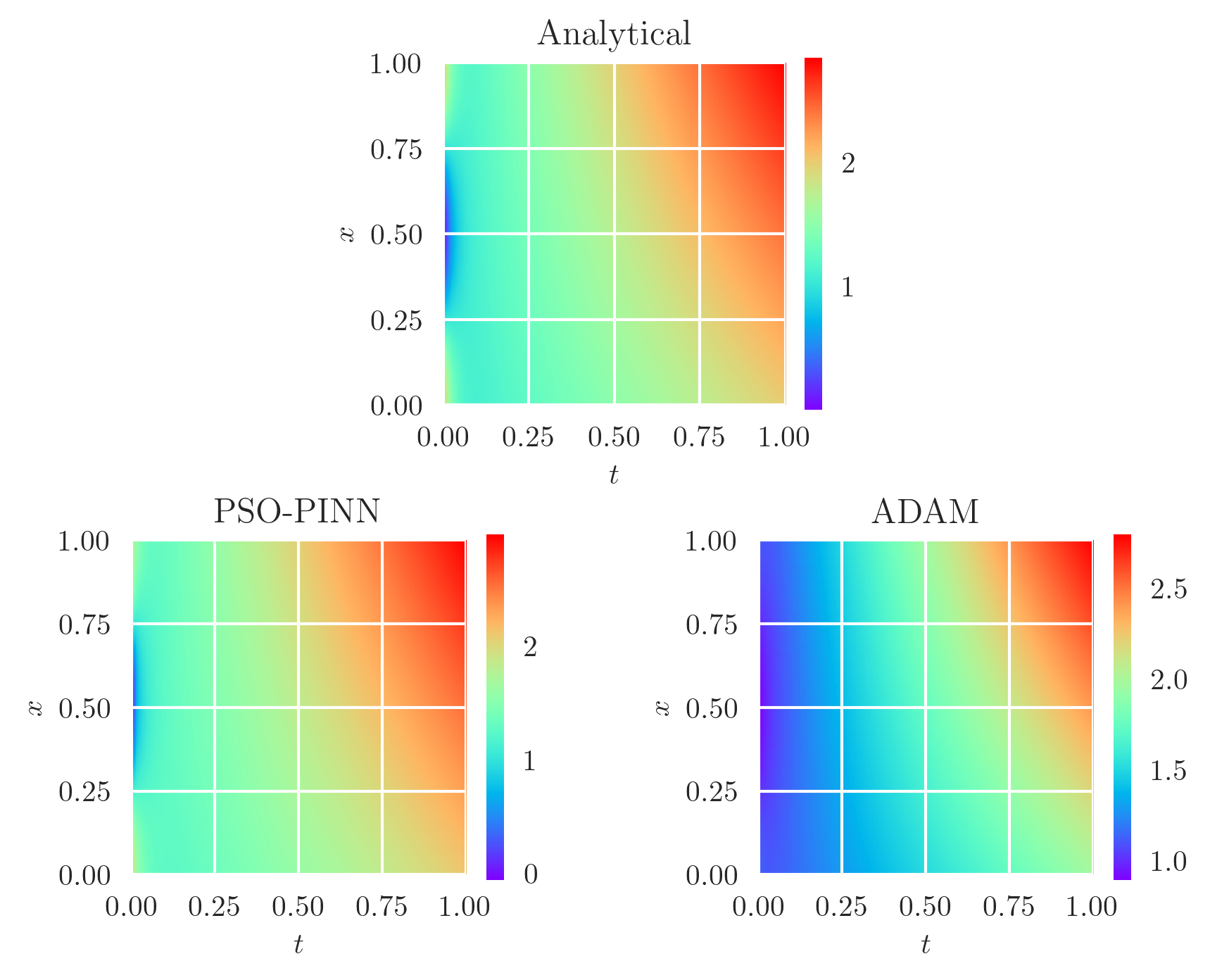} 
        \hfill
    \end{center}
    \caption{{\bf Forced Heat Equation} - Results obtained by the PSO-PINN and conventional (ADAM) PINN ensembles at the end of training.}
    \label{fig:diffusion}
\end{figure*}

In this benchmark, we set set the number of initial, boundary, and residual points to 1024, 512, and 512, respectively. We used 20 neural networks of 5 hidden layers of 8 neurons in both PSO-PINN and traditional ADAM ensembles, which are trained for 2,000 iterations. Figure \ref{fig:diffusion} displays the solution obtained by the PSO-PINN and ADAM ensembles at the end of training, where the approximation displayed is the mean of the ensemble, as before. It is clear that the PSO-PINN solution is close to the analytical solution, while the ADAM solution is rather poor. After running over 10 independent repetitions the $L_2$ error obtained by the PSO-PINN solution was $0.014\,\pm\,0.006$, while the error for the ADAM ensemble was $0.102\,\pm\,0.005$. 

\subsubsection{The Allen-Cahn Equation} 

The Allen-Cahn is a nonlinear reaction-diffusion PDE, which is used in phase-field models of the evolution of phase separation in a multi-component metal alloy. The Allen-Cahn equation considered here is:

\begin{equation}
    \label{eq:allen-cahn}
        \begin{split}
            &\frac{\partial u}{\partial t} - D\frac{\partial^2u}{\partial x^2} - 5(u - u^3) \,=\, 0, \quad x \in [-1, 1], \quad t \in [0, 1] \\
            &u(-1, t) \,=\, u(1, t)\,, \\
            &\frac{\partial u}{\partial x}\bigg\rvert_{x=-1} =\, -\frac{\partial u}{\partial x}\bigg\rvert_{x=1}\,,\\
            &u(x, 0) \,=\, x^2\cos(\pi x)\,,
        \end{split}
\end{equation}
where $D = 0.0001$ is the diffusivity coefficient. The Allen-Cahn PDE is a challenging benchmark for PINNs due to sharp space and time transitions in its solutions and the periodic boundary condition. In order to deal with the latter, the PINN boundary loss term has to be modified, as specified, e.g., in~\cite{mcclenny2020self}.

\begin{figure}
    \begin{center}
        \includegraphics[width=0.96\columnwidth]{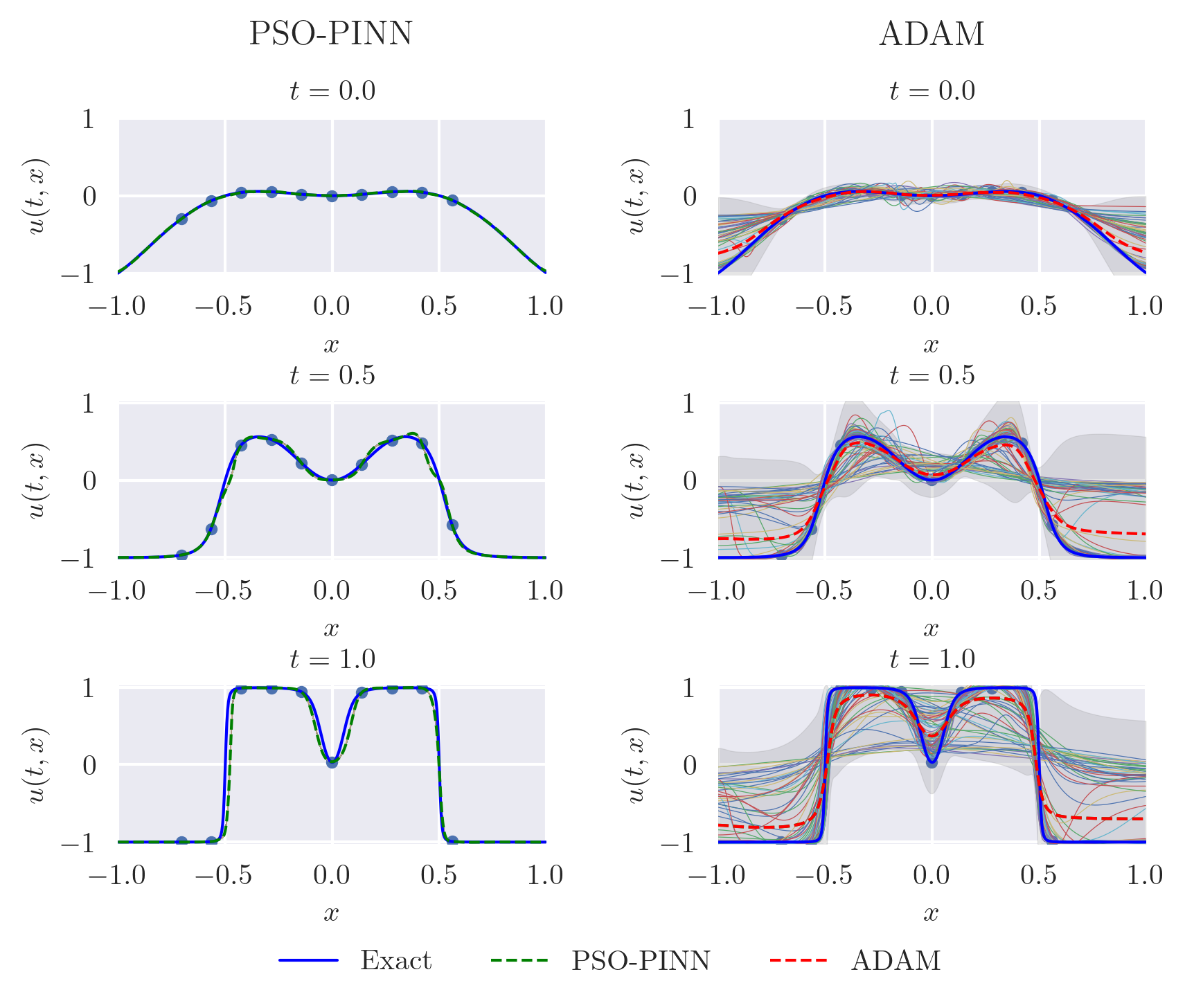} 
        \hfill
    \end{center}
\caption{{\bf Allen-Cahn Equation} - Results obtained by the PSO-PINN and conventional (ADAM) PINN ensembles at the end of training.}
\label{fig:allen-cahn}
\end{figure}

In this experiment, we set set the number of initial, boundary, and residual points to 256, 400, and 2000, respectively.  In addition to these data, we use 2000 data points randomly sampled throughout the spacial-temporal domain using latin hypercube method. We used 100 neural networks of 5 hidden layers of 8 neurons in both PSO-PINN and traditional ADAM ensembles, which are trained for 2,000 iterations.  Over 10 independent repetitions of the experiment, the PSO-PINN ensemble achieved and $L_2$ error of $0.093\,\pm\, 0.032$, while the ADAM ensemble achieved $0.327\,\pm\,0.029$. The results at the end of training for one of the repeated experiments are displayed in Figure \ref{fig:allen-cahn}. We can observe that the PSO-PINN ensemble had converged to the reference solution, while the ADAM ensemble had not. 

\subsection{Discussion}
\label{sec:discussion}

The experimental results displayed promising results for PINN training using swarm intelligence and ensemble models. It was seen in Section~4.2 that both PSO-BP and PSO-BP-CD show superior results to using a basic PSO algorithm. This is no doubt due to the extra information contained in the gradient of the function. We also observed that PSO-BP-CD, the variant we proposed in this paper, performed better than PSO-BP, especially in the case of the nonlinear Burgers benchmark.

PSO-PINN provides the advantages of an ensemble approach, allowing the use of the mean of the ensemble as the predictor and its variance as a measure of the corresponding prediction uncertainty. However, PSO-PINN has clear advantages over a standard ensemble of PINNs trained with ADAM, as seen in the results of Section 4.3. The PSO-PINN ensembles are more precise and converge faster, with well-calibrated variance, leading to a greater confidence in the solution. This no doubt results from the fact that there is communication between the PINNs in a PSO-PINN swarm, while the traditional ensemble the PINNs do not communicate. 

Setting the hyperparameters of the PSO-PINN ensemble, i.e., $\alpha$, $\beta$, $c_1$, and $c_2$, is a model selection problem comparable to setting the correct learning rates and initialization in training a deep neural neural network. Indeed, the hyper parameter $\alpha$ is directly comparable to the learning rate used in traditional gradient descent. We observed in our experimental results that suitable hyperparameter values were fairly consistent across experiments. Also, there was no need to set a schedule for the learning rate hyperparameter $\alpha$, as is usually needed in traditional gradient descent algorithms

\section{Conclusion}
\label{sec:conclusion}

In this paper, we proposed PSO-PINN, a particle-swarm optimation methodology for training physics-informed neural networks. PSO-PINN can be employed with any variant of PSO; here, we proposed PSO-BP-CD, a modification of the existing hybrid PSO-gradient descent PSO-BP method, \textcolor{black}{which puts more weight on the gradient descent component as training progresses. PSO-BP-CD was observed in our experiments to be the best-performing variant of PSO for use with PSO-PINN.} 

PSO-PINN produces an ensemble of PINN solutions, which allows variance estimation for quantifying the uncertainty in the prediction. A more considerable degree of agreement among the PINNs leads to small variance and a higher degree of confidence in the predicted values. Our experimental result indicate a clear advantage of PSO-PINN over traditional ensembles of PINNs.
Other ensemble techniques, such as stacking~\cite{wolpert1992stacked} and snapshot~\cite{huang2017snapshot}, could be readily introduced into the PSO-PINN framework to produce even better predictions and uncertainty quantification. 

As an ensemble approach, PSO-PINN incurs a larger computational cost than training a single PINN. However, some of the cost could be mitigated by using parallelization. There are out-of-the-shelf implementations in TensorFlow to parallelize tensors calculations, and PSO-PINN is already taking advantage of that. Nevertheless, there is room for improvement, mainly regarding distributed training strategies. Such procedures would allow PSO-PINN to distribute its training across multiple GPUs, clusters, multiple machines, or even Cloud TPUs. 

Some other aspects of swarm optimization are yet to be explored, such as multi-objective optimization, which would be quite suitable for PINNs, since its total loss function is nothing more than a scalarization of multiple losses (often minimized as the sum of the losses). Using distributing computing strategies will unleash further improvements in scalability and new approaches for optimization. Future work will explore these and other improvements in the PSO-PINN methodology in order to tackle more challenging problems, including inverse mapping for the determination of equation parameters and solving high-dimensional problems with unobserved boundary conditions.

\bibliographystyle{ieeetr}
\bibliography{main}

\begin{thebibliography}{10}

\bibitem{raissi2019physics}
M.~Raissi, P.~Perdikaris, and G.~E. Karniadakis, ``Physics-informed neural
  networks: A deep learning framework for solving forward and inverse problems
  involving nonlinear partial differential equations,'' {\em Journal of
  Computational Physics}, vol.~378, pp.~686--707, 2019.

\bibitem{raissi2020hidden}
M.~Raissi, A.~Yazdani, and G.~E. Karniadakis, ``Hidden fluid mechanics:
  Learning velocity and pressure fields from flow visualizations,'' {\em
  Science}, vol.~367, no.~6481, pp.~1026--1030, 2020.

\bibitem{karniadakis2021physics}
G.~E. Karniadakis, I.~G. Kevrekidis, L.~Lu, P.~Perdikaris, S.~Wang, and
  L.~Yang, ``Physics-informed machine learning,'' {\em Nature Reviews Physics},
  vol.~3, no.~6, pp.~422--440, 2021.

\bibitem{hornik1989multilayer}
K.~Hornik, M.~Stinchcombe, and H.~White, ``Multilayer feedforward networks are
  universal approximators,'' {\em Neural networks}, vol.~2, no.~5,
  pp.~359--366, 1989.

\bibitem{amari1993backpropagation}
S.-i. Amari, ``Backpropagation and stochastic gradient descent method,'' {\em
  Neurocomputing}, vol.~5, no.~4-5, pp.~185--196, 1993.

\bibitem{kingma2014adam}
D.~P. Kingma and J.~Ba, ``Adam: A method for stochastic optimization,'' {\em
  arXiv preprint arXiv:1412.6980}, 2014.

\bibitem{wang2021understanding}
S.~Wang, Y.~Teng, and P.~Perdikaris, ``Understanding and mitigating gradient
  flow pathologies in physics-informed neural networks,'' {\em SIAM Journal on
  Scientific Computing}, vol.~43, no.~5, pp.~A3055--A3081, 2021.

\bibitem{wang2022and}
S.~Wang, X.~Yu, and P.~Perdikaris, ``When and why pinns fail to train: A neural
  tangent kernel perspective,'' {\em Journal of Computational Physics},
  vol.~449, p.~110768, 2022.

\bibitem{mcclenny2020self}
L.~McClenny and U.~Braga-Neto, ``Self-adaptive physics-informed neural networks
  using a soft attention mechanism,'' {\em arXiv preprint arXiv:2009.04544},
  2020.

\bibitem{van2021optimally}
R.~van~der Meer, C.~W. Oosterlee, and A.~Borovykh, ``Optimally weighted loss
  functions for solving pdes with neural networks,'' {\em Journal of
  Computational and Applied Mathematics}, p.~113887, 2021.

\bibitem{jagtap2020extended}
A.~D. Jagtap and G.~E. Karniadakis, ``Extended physics-informed neural networks
  (xpinns): A generalized space-time domain decomposition based deep learning
  framework for nonlinear partial differential equations,'' {\em Communications
  in Computational Physics}, vol.~28, no.~5, pp.~2002--2041, 2020.

\bibitem{shukla2021parallel}
K.~Shukla, A.~D. Jagtap, and G.~E. Karniadakis, ``Parallel physics-informed
  neural networks via domain decomposition,'' {\em arXiv preprint
  arXiv:2104.10013}, 2021.

\bibitem{kennedy1995particle}
J.~Kennedy and R.~Eberhart, ``Particle swarm optimization,'' in {\em
  Proceedings of ICNN'95-international conference on neural networks}, vol.~4,
  pp.~1942--1948, IEEE, 1995.

\bibitem{chakraborty2017swarm}
A.~Chakraborty and A.~K. Kar, ``Swarm intelligence: A review of algorithms,''
  {\em Nature-Inspired Computing and Optimization}, pp.~475--494, 2017.

\bibitem{van2000cooperative}
F.~Van~den Bergh and A.~P. Engelbrecht, ``Cooperative learning in neural
  networks using particle swarm optimizers,'' {\em South African Computer
  Journal}, vol.~2000, no.~26, pp.~84--90, 2000.

\bibitem{das2014artificial}
G.~Das, P.~K. Pattnaik, and S.~K. Padhy, ``Artificial neural network trained by
  particle swarm optimization for non-linear channel equalization,'' {\em
  Expert Systems with Applications}, vol.~41, no.~7, pp.~3491--3496, 2014.

\bibitem{mirjalili2015effective}
S.~Mirjalili, ``How effective is the grey wolf optimizer in training
  multi-layer perceptrons,'' {\em Applied Intelligence}, vol.~43, no.~1,
  pp.~150--161, 2015.

\bibitem{mousavirad2020benchmark}
S.~J. Mousavirad, G.~Schaefer, S.~M.~J. Jalali, and I.~Korovin, ``A benchmark
  of recent population-based metaheuristic algorithms for multi-layer neural
  network training,'' in {\em Proceedings of the 2020 genetic and evolutionary
  computation conference companion}, pp.~1402--1408, 2020.

\bibitem{hansen1990neural}
L.~K. Hansen and P.~Salamon, ``Neural network ensembles,'' {\em IEEE
  transactions on pattern analysis and machine intelligence}, vol.~12, no.~10,
  pp.~993--1001, 1990.

\bibitem{krizhevsky2012imagenet}
A.~Krizhevsky, I.~Sutskever, and G.~E. Hinton, ``Imagenet classification with
  deep convolutional neural networks,'' {\em Advances in neural information
  processing systems}, vol.~25, pp.~1097--1105, 2012.

\bibitem{lakshminarayanan2016simple}
B.~Lakshminarayanan, A.~Pritzel, and C.~Blundell, ``Simple and scalable
  predictive uncertainty estimation using deep ensembles,'' {\em arXiv preprint
  arXiv:1612.01474}, 2016.

\bibitem{deng2014ensemble}
L.~Deng and J.~Platt, ``Ensemble deep learning for speech recognition,'' in
  {\em Proc. interspeech}, 2014.

\bibitem{qiu2014ensemble}
X.~Qiu, L.~Zhang, Y.~Ren, P.~N. Suganthan, and G.~Amaratunga, ``Ensemble deep
  learning for regression and time series forecasting,'' in {\em 2014 IEEE
  symposium on computational intelligence in ensemble learning (CIEL)},
  pp.~1--6, IEEE, 2014.

\bibitem{cao2020ensemble}
Y.~Cao, T.~A. Geddes, J.~Y.~H. Yang, and P.~Yang, ``Ensemble deep learning in
  bioinformatics,'' {\em Nature Machine Intelligence}, vol.~2, no.~9,
  pp.~500--508, 2020.

\bibitem{ganaie2021ensemble}
M.~Ganaie, M.~Hu, {\em et~al.}, ``Ensemble deep learning: A review,'' {\em
  arXiv preprint arXiv:2104.02395}, 2021.

\bibitem{haitsiukevich2022improved}
K.~Haitsiukevich and A.~Ilin, ``Improved training of physics-informed neural
  networks with model ensembles,'' {\em arXiv preprint arXiv:2204.05108}, 2022.

\bibitem{yadav2019ga}
R.~K. Yadav {\em et~al.}, ``Ga and pso hybrid algorithm for ann training with
  application in medical diagnosis,'' in {\em 2019 Third International
  Conference on Intelligent Computing in Data Sciences (ICDS)}, pp.~1--5, IEEE,
  2019.

\bibitem{lecun2015deep}
Y.~LeCun, Y.~Bengio, and G.~Hinton, ``Deep learning,'' {\em nature}, vol.~521,
  no.~7553, pp.~436--444, 2015.

\bibitem{glorot2011deep}
X.~Glorot, A.~Bordes, and Y.~Bengio, ``Deep sparse rectifier neural networks,''
  in {\em Proceedings of the fourteenth international conference on artificial
  intelligence and statistics}, pp.~315--323, JMLR Workshop and Conference
  Proceedings, 2011.

\bibitem{baydin2018automatic}
A.~G. Baydin, B.~A. Pearlmutter, A.~A. Radul, and J.~M. Siskind, ``Automatic
  differentiation in machine learning: a survey,'' {\em Journal of machine
  learning research}, vol.~18, 2018.

\bibitem{eberhart1995particle}
R.~Eberhart and J.~Kennedy, ``Particle swarm optimization,'' in {\em
  Proceedings of the IEEE international conference on neural networks}, vol.~4,
  pp.~1942--1948, Citeseer, 1995.

\bibitem{shi1999empirical}
Y.~Shi and R.~C. Eberhart, ``Empirical study of particle swarm optimization,''
  in {\em Proceedings of the 1999 congress on evolutionary computation-CEC99
  (Cat. No. 99TH8406)}, vol.~3, pp.~1945--1950, IEEE, 1999.

\bibitem{wang2018particle}
D.~Wang, D.~Tan, and L.~Liu, ``Particle swarm optimization algorithm: an
  overview,'' {\em Soft Computing}, vol.~22, no.~2, pp.~387--408, 2018.

\bibitem{glorot2010understanding}
X.~Glorot and Y.~Bengio, ``Understanding the difficulty of training deep
  feedforward neural networks,'' in {\em Proceedings of the thirteenth
  international conference on artificial intelligence and statistics},
  pp.~249--256, JMLR Workshop and Conference Proceedings, 2010.

\bibitem{tensorflow2015-whitepaper}
M.~Abadi, A.~Agarwal, P.~Barham, E.~Brevdo, Z.~Chen, C.~Citro, G.~S. Corrado,
  A.~Davis, J.~Dean, M.~Devin, S.~Ghemawat, I.~Goodfellow, A.~Harp, G.~Irving,
  M.~Isard, Y.~Jia, R.~Jozefowicz, L.~Kaiser, M.~Kudlur, J.~Levenberg,
  D.~Man\'{e}, R.~Monga, S.~Moore, D.~Murray, C.~Olah, M.~Schuster, J.~Shlens,
  B.~Steiner, I.~Sutskever, K.~Talwar, P.~Tucker, V.~Vanhoucke, V.~Vasudevan,
  F.~Vi\'{e}gas, O.~Vinyals, P.~Warden, M.~Wattenberg, M.~Wicke, Y.~Yu, and
  X.~Zheng, ``{TensorFlow}: Large-scale machine learning on heterogeneous
  systems,'' 2015.
\newblock Software available from tensorflow.org.

\bibitem{leveque2002finite}
R.~J. LeVeque {\em et~al.}, {\em Finite volume methods for hyperbolic
  problems}, vol.~31.
\newblock Cambridge university press, 2002.

\bibitem{wolpert1992stacked}
D.~H. Wolpert, ``Stacked generalization,'' {\em Neural networks}, vol.~5,
  no.~2, pp.~241--259, 1992.

\bibitem{huang2017snapshot}
G.~Huang, Y.~Li, G.~Pleiss, Z.~Liu, J.~E. Hopcroft, and K.~Q. Weinberger,
  ``Snapshot ensembles: Train 1, get m for free,'' {\em arXiv preprint
  arXiv:1704.00109}, 2017.

\end{thebibliography}

\end{document}